\documentclass{article}
\usepackage{spconf,amsmath, multirow, tabulary,verbatim} 
\usepackage{graphicx}
\usepackage{subfig}
\usepackage{xcolor}
\usepackage{bbding}
\usepackage{url}

\title{Exploring State-of-the-art models for Early Detection of Forest Fires}
\name{Sharjeel Ahmed, Daim Armaghan, Fatima Naweed, Umair Yousaf, Ahmad Zubair and Murtaza Taj}
\address{Computer Vision and Graphics Lab, 
 LUMS School of Science and Engineering}
\begin{document}
%
\maketitle
\begin{abstract}

There have been many recent developments in the use of Deep Learning Neural Networks for fire detection. In this paper, we explore an early warning system for detection of forest fires. Due to the lack of sizeable datasets and models tuned for this task, existing methods suffer from missed detection. In this work, we first propose a dataset for early identification of forest fires through visual analysis. Unlike existing image corpuses that contain images of wide-spread fire, our dataset consists of multiple instances of smoke plumes and fire that indicates the initiation of fire. We obtained this dataset synthetically by utilising game simulators such as Red Dead Redemption 2. We also combined our dataset with already published images to obtain a more comprehensive set. Finally, we compared image classification and localisation methods on the proposed dataset. More specifically we used YOLOv7 (You Only Look Once) and different models of detection transformer.

\end{abstract}


%
\section{Introduction}
\label{sec:intro}
\label{ssec:subhead1}
Wildfires are a substantial economical and ecological threat, that require large scale strategies and funding to prevent. The problem has increased manifolds during the last few decades due to global warming. According to an estimate the earth is getting warmer at a rate of 0.32° F (0.18° C) per decade~\cite{NOAAClimate2022}, which is leading towards a $50\%$ increase in wild fire incidents. Other common reasons behind forest fires are human negligence, arson or burning for agriculture. Wildfires due to climate change can be predicted by building fuel models~\cite{RothermelVacchiano2014} and analysing remote sensing and weather station data~\cite{Yang_Lupascu_Meel_2021}. However man-made forest fires require continuous ground-based monitoring of the forest area which is typically performed by forest watch towers. These towers are equipped with various imaging sensors such as static cameras, pan-tilt and zoom cameras (referred to as PTZ cameras) and thermal cameras that can monitor an area of up to a 15 $km$ radius. However, in order to make them effective, automated image processing is required.

Image-based fire detection algorithms use live image data from the installed cameras to detect fires~\cite{FoggiaTCSVT2015, PengCEA2019}. They typically have three main stages: data pre-processing, feature extraction and detection. In traditional algorithms, feature selection is done manually using professional knowledge which makes it hard to extract complex features and results in lower accuracy and weak generalization ability~\cite{FoggiaTCSVT2015}. Image recognition algorithms based on convolutional neural networks (CNNs) can automatically learn and extract complex image features effectively~\cite{ZHOUFire2016, ThermalImagigPuLi2020, Analysiskukuk_comprehensive_2021, RobertCVBC2021fire, Saponara2021real, QingMDPI2022}. 

\begin{figure}
    \centering
    \begin{tabular}{cccc}
    \subfloat[]{\includegraphics[width=.2\columnwidth]{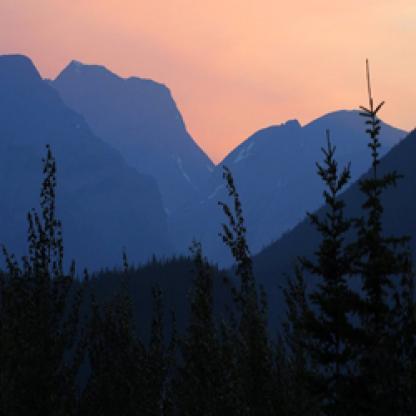}} &
    \subfloat[]{\includegraphics[width=.2\columnwidth]
    {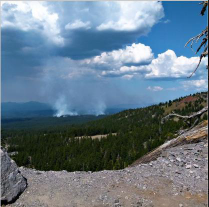}} &
    \subfloat[]{\includegraphics[width=.2\columnwidth]{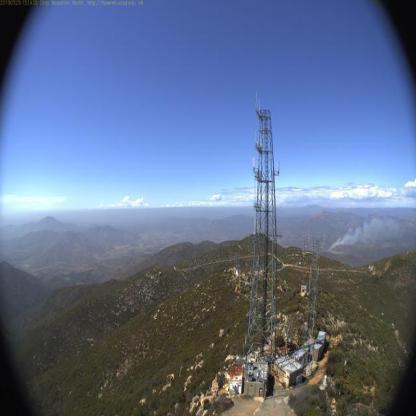}} &
    \subfloat[]{\includegraphics[width=.2\columnwidth]{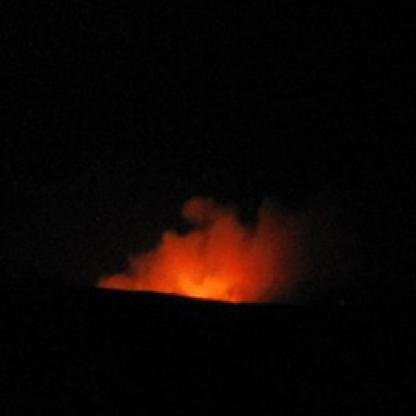}}
    \end{tabular}
    \vspace{-.2cm}
    \caption{Challenges in forest fire detection. (a) Orange colorization of the sky at sunset. (b) Ambiguity between smoke, clouds and fog. (c) Small smoke plume at a very distant location. (d) Identifying smoke at night without thermal imaging. }
    \label{fig:challenges}
    \vspace{-.5cm}
\end{figure}





Algorithms for automatic identification of fire or smoke in the images can be categorized into image classification~\cite{RobertCVBC2021fire},  segmentation~\cite{ZHOUFire2016} and localization~\cite{ThermalImagigPuLi2020, Saponara2021real, QingMDPI2022}.  SmokiFi-CNN~\cite{RobertCVBC2021fire} is an image classification based technique, that tends to classify images into either normal, smoke, fire or fire with smoke classes. During pre-processing, segmentation between the background and foreground is performed, along with white balancing to further highlight areas with smoke which helps in reducing the number of missed detections. The flames of wildfires can almost be invisible from long distances. However, the rising smoke plumes generated by the fire can usually be seen in the viewing field of the camera. Therefore, it would be more practical to detect the smoke plumes instead of detecting flames for the purpose of long-distance wildfire detection. A wildfire smoke detection method based on local extremal region segmentation~\cite{ZHOUFire2016} was proposed to localise smoke pixels. This method used a linear time Maximally Stable Extremal Regions (MSERs)\cite{MSERmatas2004robust} detection algorithm to make the initial segmentation less dependent on motion and color information.

\begin{figure}[t]
    \centering
    \subfloat[Smoke Augmented]{\includegraphics[width=.2\columnwidth]{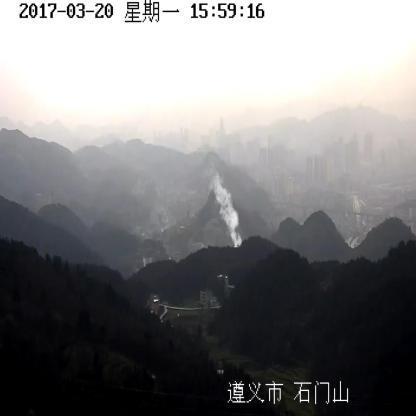}} \hfill
    \subfloat[Fire]{\includegraphics[width=.2\columnwidth]{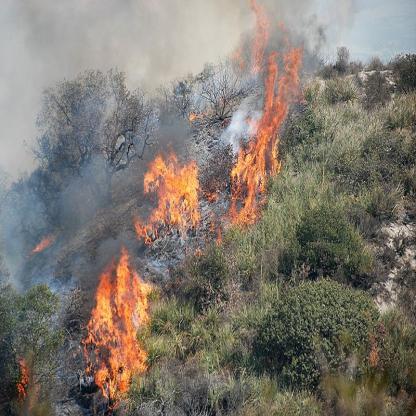}} \hfill
    \subfloat[Fire]{\includegraphics[width=.2\columnwidth]{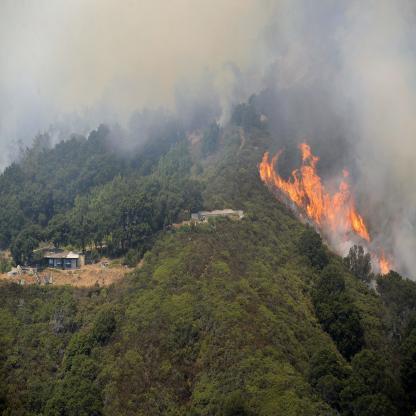}} \hfill
    \subfloat[Smoke Synthetic]{\includegraphics[width=.2\columnwidth]{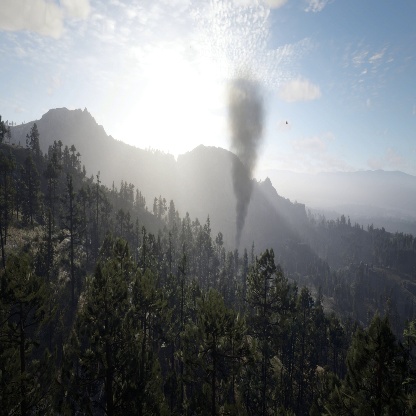}}\\
    \subfloat[Smoke]{\includegraphics[width=.2\columnwidth]{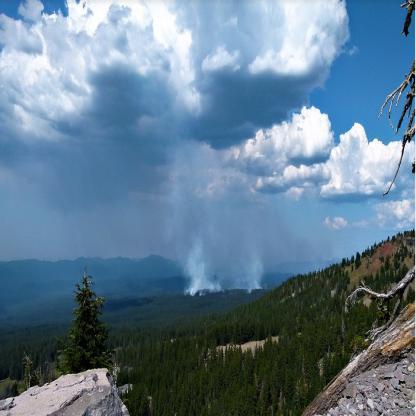}} \hfill
    \subfloat[Fire]{\includegraphics[width=.2\columnwidth]{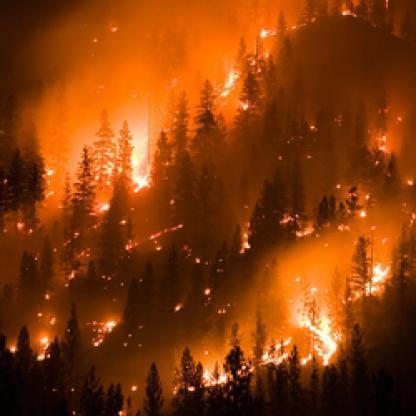}} \hfill
    \subfloat[Fire Synthetic]{\includegraphics[width=.2\columnwidth]{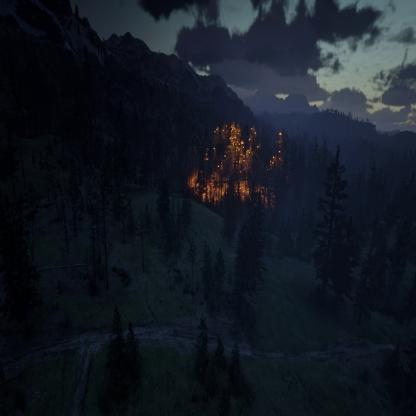}} \hfill
    \subfloat[Smoke]{\includegraphics[width=.2\columnwidth]{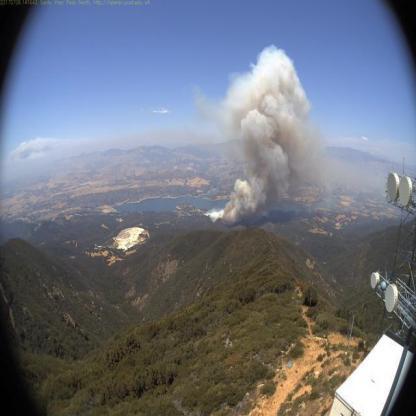}}
    \caption{Sample images from dataset including samples of both synthetic, augmented and real images.}
    \label{fig:SOAdata}
    \vspace{-0.5cm}
\end{figure}

Most of the localization algorithms perform both classification as well as extraction of bounding boxes. These include object detectors such as variants of Region CNN~\cite{ren2015faster}, YOLO~\cite{JosephYoloCVPR2017} and detection transformer~\cite{DETRECCV2022}. A comparison of these methods for the application of forest fire detection has been performed which reveals that YOLOv3 achieved better accuracy at low compute cost~\cite{ThermalImagigPuLi2020}. Many of the recent methods are thus based on variants of YOLO. Since a practical forest fire surveillance system requires continuous monitoring of the site via a live camera feed, thus real-time performance is usually considered an important aspect~\cite{Saponara2021real}. This work focuses on both outdoor and indoor detection. Simultaneously, it reduces the need for storage and compute via YOLOv2, which has a less complex model and reduced number of weights. This allows its deployment on smaller devices, while providing results within 2 seconds, which is almost 25 times faster than the R-CNN~\cite{ren2015faster}. Similarly, a variant of YOLOv5 is used on localisation of fire in both ground and aerial imagery~\cite{QingMDPI2022}. This method attempts to reduce the error rate by improving the anchor box clustering via a variant of K-means algorithm. It also prunes the network head of YOLOv5 to further improve the detection speed. A more comprehensive survey of fire detection can be found in~\cite{Analysiskukuk_comprehensive_2021}.

A major short coming of existing methods is, although they detect wide-spread wild fires in the vicinity of camera,  they are inapplicable for early detection of forest fires via identification of smoke plumes. Major reasons include lack of sizeable datasets of smoke from the initial stages of fire, inherent ambiguities between smoke, cloud, fog, smog as well as changes in sky appearances at sunset and sunrise (see Fig.~\ref{fig:challenges}).

We first propose a dataset for early identification of forest fires through visual analysis. Unlike, existing image corpuses that contains images of wide-spread fire, our dataset consists of multiple instances of smoke plumes and fire that indicate the initiation of fire. We obtained this dataset synthetically via game simulator Red Dead Redemption 2~\cite{ReddeadRedemption2}. We also combined our dataset with already published images to obtain a more comprehensive set. Finally, we compared image classification and localisation methods on the proposed dataset. More specifically, we used YOLOv7 (You Only Look Once)~\cite{YOLOv7} and different models of detection transformer~\cite{DETRECCV2022}. 


\begin{table}[t]
    \centering
    \begin{tabular}{ccc}
        \multirow{2}{*}{\includegraphics[width=.3\columnwidth]{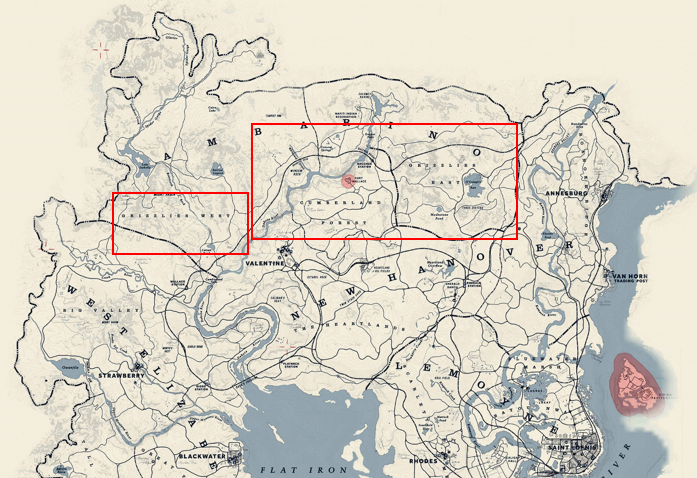}} & 
        \includegraphics[width=.2\columnwidth]{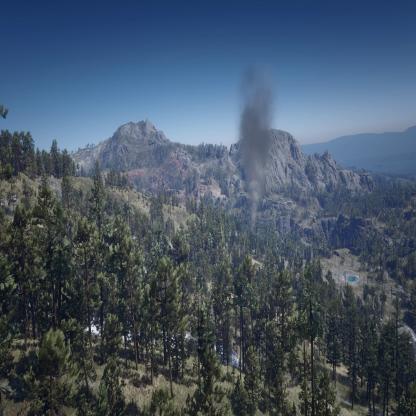} &
        \includegraphics[width=.2\columnwidth]{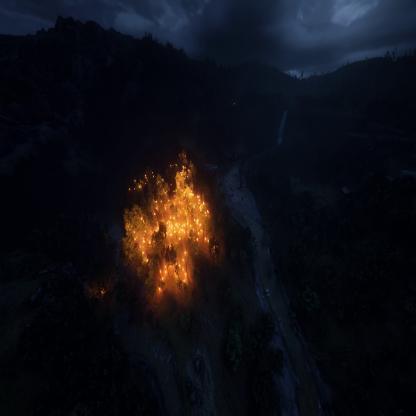} \\ 
        & \includegraphics[width=.2\columnwidth]{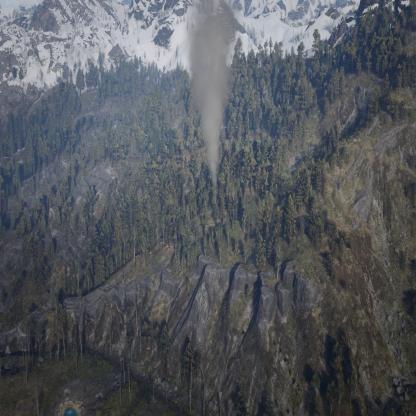} &
        \includegraphics[width=.2\columnwidth]{images/dataset_img7.jpg} \\
        (a) & (b) & (c)\\
    \end{tabular}
    \caption{Synthetic dataset samples from Red Dead Redemption 2. (a) Map image showing forest region. \cite{rdr2_map} (b-c) Sample instance of smoke and fire events.}
    \label{tab:reddeadmap}
    \vspace{-.5cm}
\end{table}

\section{Dataset}
Although there are many cameras installed around the world to watch forest areas, since fire is a rare event, most of the existing datasets from real forest watch towers contain vast majority of normal images (i.e. image without any fire or smoke). While the other datasets contain images from intense fire, or fire and smoke augmented in the scene (see Table~\ref{tab:dataset} \& Fig.~\ref{fig:SOAdata}). 

Thus a major challenge is to find a sizeable, well balanced dataset to train our models. We addressed this problem by generating synthetic data from 3 main regions, Cumberland Forest, Grizzlies East and Grizzlies West (see Table~\ref{tab:reddeadmap}) which are regions in the computer game titled Red Dead Redemption 2~\cite{ReddeadRedemption2}. The game features meticulous, true to life graphics. These 3 regions were selected due to there being several mountains and extensive vegetation and trees. Mods were used to place smoke within the trees to simulate a fire in its initial stages, as well as to place a large number of campfires to simulate a fire that had been burning for some time. Screen recording of these scenarios from various angles and focal lengths were done. Frames were extracted from these recordings and a synthetic dataset was generated for \emph{smoke} and \emph{fire} classes. We combined this with other resources to form our final dataset.

\begin{table}[b]
    \centering
    \scalebox{.85}
    {
    \begin{tabular}
    { 
|p{0.15\linewidth}|p{0.08\linewidth}|p{0.08\linewidth}|p{0.08\linewidth}|p{0.08\linewidth}|p{0.08\linewidth}|p{0.08\linewidth}|}
     \hline
      & \multicolumn{2}{c|}{\bf{Train}} &\multicolumn{2}{c|}{\bf{Valid}} & \multicolumn{2}{c|}{\bf{Test}} \\
     \hline
     Class & Im. & Inst. & Im. & Inst. & Im. & Inst. \\
     \hline
     Smoke  & 1257 & 1448 & 403 & 467 & 378 & 432\\
     \hline
     Fire & 1592 & 2657 & 532 & 909 & 497 & 808\\
     \hline
     Normal  & 2656 &2656 & 907 &907 & 937 & 937\\
     \hline
    \end{tabular}
    }
    \captionof{table}{Train-Validation-Test split in dataset along each class (Im.: Images, Inst.: Instances).}
\end{table}

\begin{table}[h]
    \resizebox{\columnwidth}{!}{
    \setlength\extrarowheight{2pt}
    \begin{tabular}{
    |p{0.3\linewidth}|p{0.08\linewidth}|p{0.1\linewidth}|p{0.1\linewidth}|
    p{0.1\linewidth}|p{0.1\linewidth}|p{0.08\linewidth}|p{0.08\linewidth}|
    }
    \hline
         & \bf{Ann.} & \bf{Imgs} & \bf{Inst.} & \bf{R/Sy/A} & \bf{Views} & \bf{Type} & \bf{Used} \\ \hline
        Mivia (Fire)\cite{mivia} & Class & Videos & 106 & R & 8 & S & Y \\ \hline
        Mivia (Smoke)\cite{mivia_smoke} & Class & Videos & 300 & R & 1 & S & Y \\ \hline
        Kaggle 1\cite{alik05_2022} & Class & 1900 & 1900 & R & Mult & S & Y \\ \hline
        Kaggle 2\cite{kutlu_2021} & Class & 15734 & 15734 & Aug & Mult & S & Y \\ \hline
        Mendeley \cite{khan_2020} & Class & 1900 & 1900 & R & Mult & S & N \\ \hline
        Images.cv \cite{imagescvdataset} & Class & 948 & 948 & R & Mult & S & Y \\ \hline
        HPWREN \cite{hpwren} & None & Videos & - & R & 50+ & S & Y \\ \hline
        RoboFlow \cite{mankind_2022} & BB & 737 & 737 & Aug & Few & FE & N \\ \hline
        NIST\cite{nist_2021} & Class & Videos & - & R & Mult & 360 & N \\ \hline
        HPWREN\cite{aiformankind} & BB & 2192 & 2192 & R & Mult & FE & Y \\ \hline
        Read Dead 2\cite{ReddeadRedemption2} & Class & 1531 & 1531 & Synth & Mult & S & Y \\ \hline
    \end{tabular}
    }
    \captionof{table}{Summary of existing and generated datasets for fire and smoke detection (Ann.: Annotations, Class: Classification, BB: Bounding Box, Imgs.: Images, R: Real, Sy: Synthetic: A: Augmented, S: Street view, FE: Fish Eye, Mult: Multiple).}
    \label{tab:dataset}
\end{table}

\begin{table}[t]
    \centering
    \scalebox{0.85}
    {
\begin{tabular}{ |p{1.3cm}|p{1.2cm}|p{1.2cm}|p{1.7cm}|p{1.2cm}|c|  }
 \hline
 | & \bf{YOLOv7} & \bf{YOLOv7 -tiny} & \bf{Def. DETR} & \bf{DETR} \\
 \hline 
 GPU & Tesla T4 & Tesla T4 & Tesla T4 & Tesla T4 \\
 \hline
 LR & 0.001 & 0.001 & 0.002 & 0.001\\
 \hline
 Epochs & 50 & 50 & 37 & 65 \\
 \hline
 Batch & 8 & 8 & 4 & 4 \\
 \hline
 CT & 0.1 & 0.1 & 0.5 & 0.5 \\
 \hline
\end{tabular}
}
\captionof{table}{Summary of training and testing conditions for the four models. (LR: Learning rate, Batch: the batch size of each iteration, CT: Confidence Threshold).
} 
    \label{tab:traintest}
\end{table}


\section{Models \& Classifier}
\label{ssec:subhead}
We selected four state-of-the-art object detectors in this work. We for the first time introduced transformer based networks such as Detector Transformer (DETR)~\cite{DETRECCV2022} and its variant Deformable DETR~\cite{DefDETEICLR2021} and compared their performance with variants of YOLO (YOLOv7~\cite{YOLOv7}) and YOLOv7-tiny).

{\textbf{YOLOv7}}: YOLO-v7~\cite{YOLOv7} belongs to a series of state-of-the-art models, YOLO (You Only Look Once). As opposed to other object detectors which focus on specified regions of an image, YOLO views the complete image at once. YOLO processes images faster than most object detectors, and is therefore commonly used for real-time detections. \\
Fine-tuning was performed on pre-trained YOLO-v7. We trained the model weights for 50 epochs with a batch size of 8, and learning rate of 0.01. It utilizes an SGD optimizer with Nestrov momentum for back propagation and SiLU as its activation function. 

{\textbf{YOLOv7-tiny}}: It is a version of YOLO-v7 optimized for edge GPU computation. Compared to the YOLO-v7 standard model, it only uses $16\%$ of the number of parameters (See Table ~\ref{tab:traintest}). Consequently, it leads to faster inference and can run on lightweight devices that are much more suitable for real-time deployment. \\
Fine-tuning was performed on pre-trained YOLO-v7-tiny model weights. We trained the model weights for 50 epochs with a batch size of 8, and learning rate of 0.01. It utilizes an SGD optimizer with Nestrov momentum for back propagation and ReLU as its activation function.

{\textbf{{DETR}}: DETR (Detection Transformer)~\cite{DETRECCV2022} views object detection as a direct set prediction problem, incorporating a set-based global loss that forces unique predictions via bi-partite matching and a transformer encoder-decoder architecture. We used the DETR pre-trained model and fine-tuned it on our dataset without making any changes to their hyper-parameters.

{\textbf{Deformable DETR}}: Deformable DETR\cite{DefDETEICLR2021} is based on DETR and aims to speed up the model's convergence and increase feature spatial resolution caused by the
limitation of Transformer attention modules in processing image feature maps.
Deformable DETR achieves better performance than DETR with fewer training epochs. We used the Deformable DETR (single scale) pre-trained model and fine-tuned it on our dataset without making any changes to their hyper-parameters.   



\section{Results and Conclusion}
\label{sec:results}

    
\begin{table}[t]
    \centering
        \scalebox{0.8}
        {
        \begin{tabular}{ |p{2.3cm}|p{1.2cm}|p{1.2cm}|p{1.7cm}|p{1.2cm}|c|  }
             \hline
             | & \bf{YOLOv7} & \bf{YOLOv7-tiny} & \bf{Def. DETR} & \bf{DETR} \\
             \hline
            Accuracy & 0.65 & 0.88 & 0.81 & 0.62 \\
             \hline
            Precision & 0.72 & 0.89 & 0.94 & 1.0 \\
             \hline
            Recall & 0.67 & 0.89 & 0.65 & 0.17 \\
             \hline
            F1 & 0.64 & 0.88 & 0.74 & 0.29 \\
             \hline
             \hline
            mAP@ 0.5 & 0.40 & 0.41  & 0.46 & 0.36 \\
             \hline
            mAP@ 0.5:0.95 & 0.17 & 0.17 & 0.21 & 0.13 \\
             \hline
             \hline
            Params (M) & 37.20  & 6.02 & 40.80 & 41.28 \\
             \hline
            Inf. time & 0.15 & 0.2 & 2.1 & 2.4 \\             
            \hline                
        \end{tabular}
        }
        \captionof{table}{Comparison of different performance metrics of the four different models. (Row 1-4) Classification results (Inf. time: Inference time for one image in milliseconds(ms), M: 1 million parameters).} \label{tab:performance} 
\end{table}

    \begin{figure}[b]
        \scalebox{0.85}{
            \subfloat[YOLOv7]{
                \includegraphics[width=2.2 cm, height=2.2 cm]
                {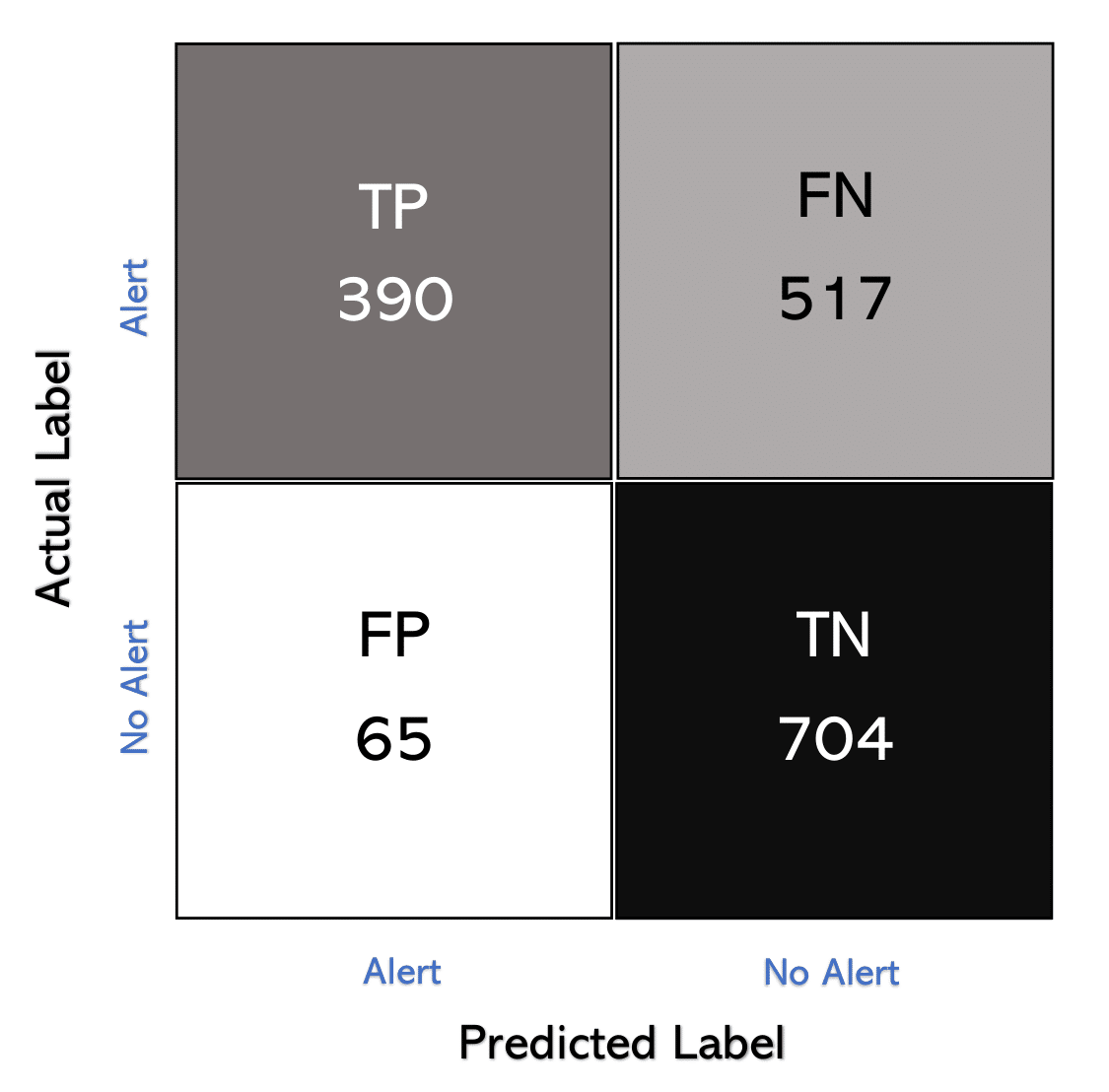}
            }
            \subfloat[YOLOv7-tiny]{
                \includegraphics[width=2.2 cm, height=2.2 cm]
                {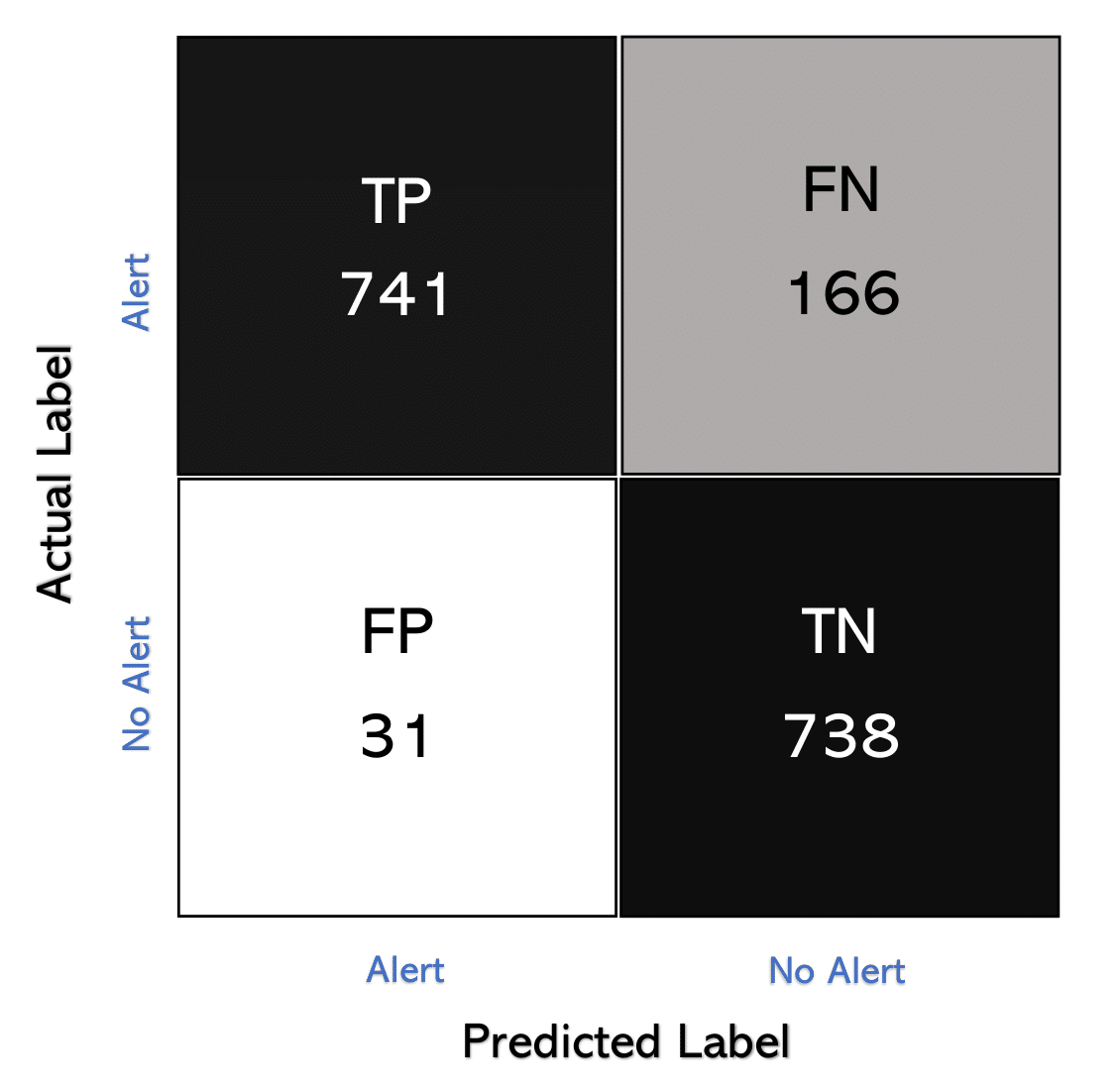} 
            } \\
            \subfloat[Deformable DETR]{
                \includegraphics[width=2.2 cm, height=2.2 cm]
                {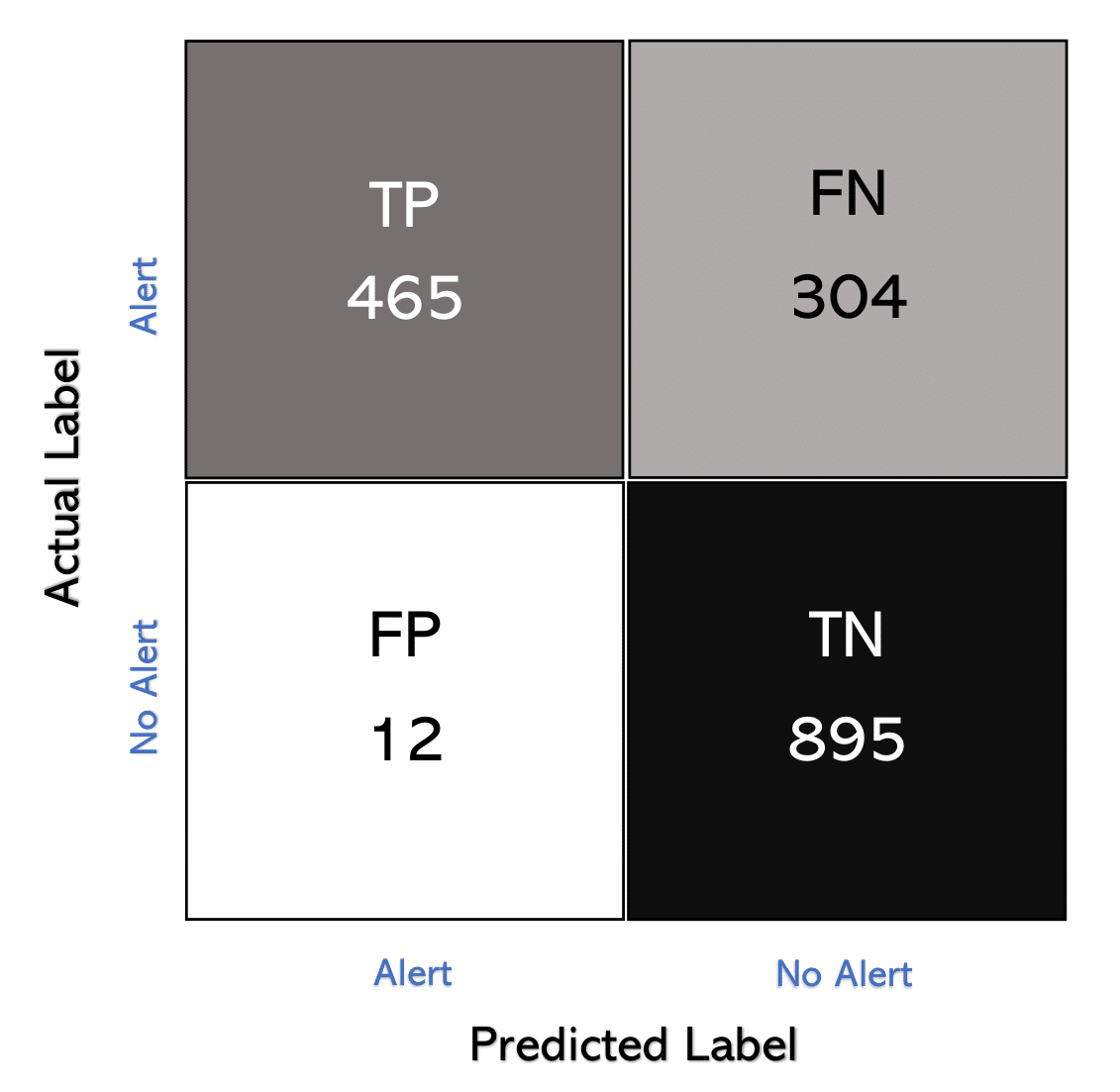}
            }
            \subfloat[DETR]{
                \includegraphics[width=2.2 cm, height=2.2 cm]
                {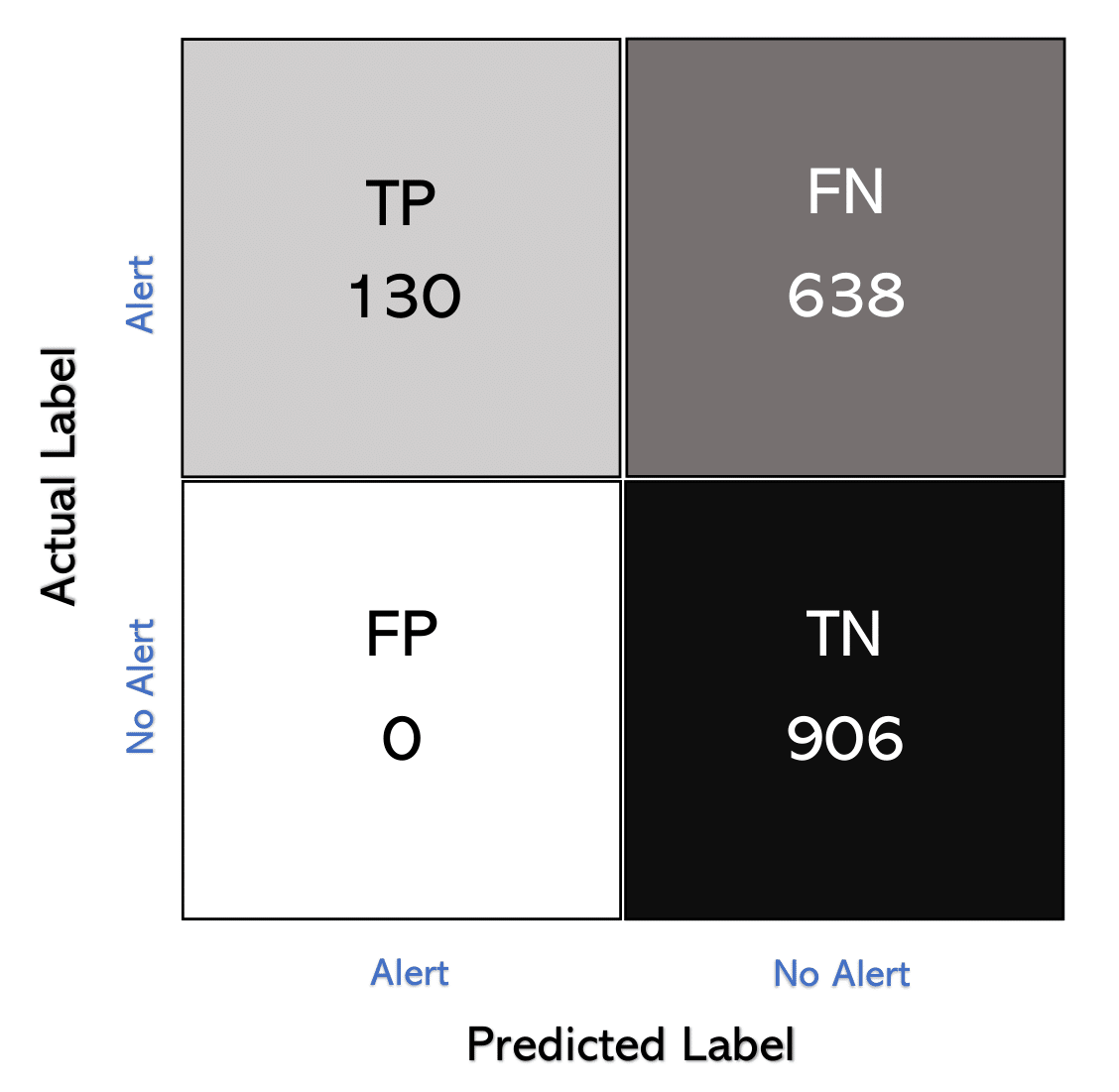} 
            } 
        }
        \caption{Confusion matrices for the four models show us the number of True and False Positives and True and False Negatives.}
        \label{fig:conf}
    \end{figure}
        
    \begin{figure}[ht]
        \begin{tabular}
        {p{0.2\linewidth}|p{0.07\linewidth}|p{0.13\linewidth}|p{0.13\linewidth}| p{0.1\linewidth}| p{0.1\linewidth}| }
         \cline{2-6}
          & GT & YOLO v7 & YOLO v7-tiny & Def. DETR & DETR  \\
         \cline{2-6}
        \includegraphics[width=1.5 cm, height=1.5 cm]{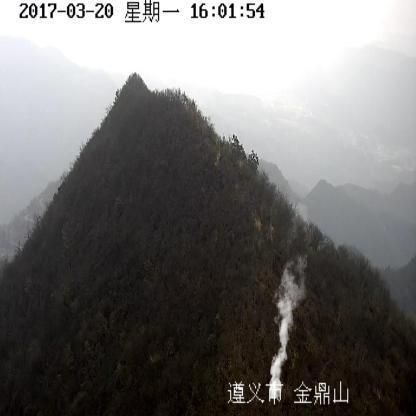} &
         S  & S & S & S & N \\
        \cline{2-6}
        \includegraphics[width=1.5 cm, height=1.5 cm]{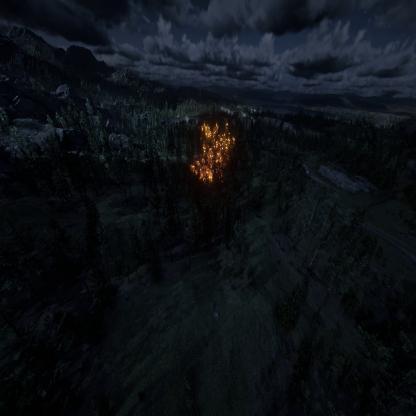} &
        F & F & F & F & F  \\
        \cline{2-6}
        \includegraphics[width=1.5 cm, height=1.5 cm]{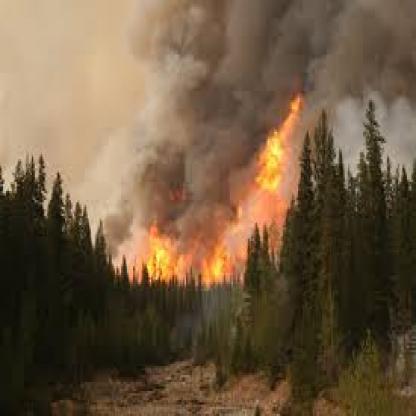} &
        F/S & F & F & F & N\\
        \cline{2-6}
        \includegraphics[width=1.5 cm, height=1.5 cm]{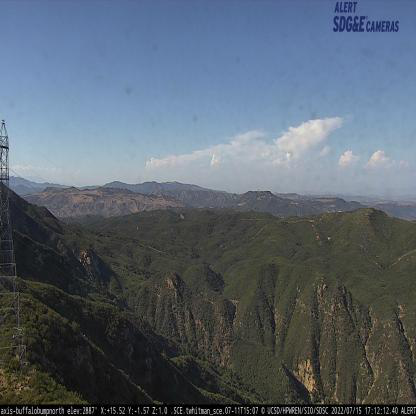} &
         N & F/S & N & N & N\\
         \cline{2-6}
        \end{tabular}
    \caption{Inference results for classification. (GT: Ground truth, S: Smoke, F: Fire, N: Normal).}
    \vspace{-0.5cm}
    \end{figure}
    
    \begin{figure}[ht]
        \begin{tabular}{ 
            p{0.2\linewidth}p{0.2\linewidth}p{0.2\linewidth}p{0.2\linewidth}p{0.2\linewidth}  }
            GT & YOLOv7 & YOLOv7-tiny & Def. DETR & DETR 
            \\
            \includegraphics[width=1.5 cm, height=1.5 cm]{sample_images/originals/001182_jpg.rf.18ff5097f38c00007b4e559fd8838133.jpg} &
            \includegraphics[width=1.5 cm, height=1.5 cm]
            {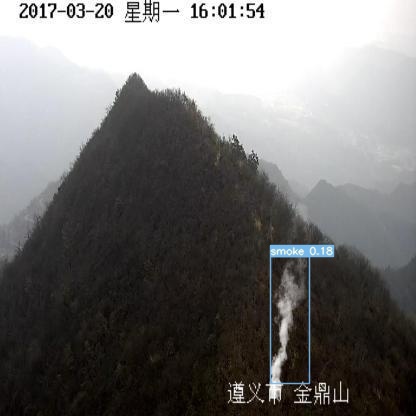} &
            \includegraphics[width=1.5 cm, height=1.5 cm]
            {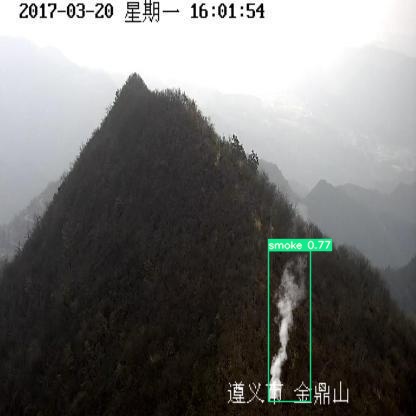} &
            \includegraphics[width=1.5 cm, height=1.5 cm]
            {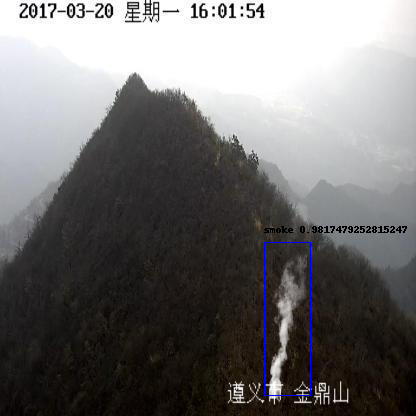} &
            \includegraphics[width=1.5 cm, height=1.5 cm]
            {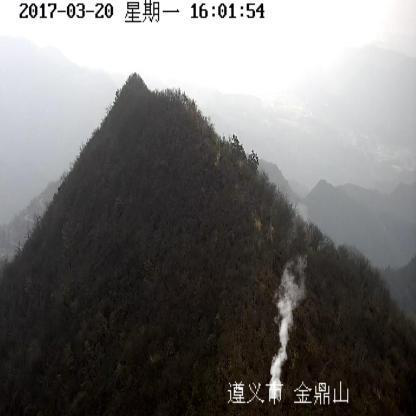} 
            \\
            \includegraphics[width=1.5 cm, height=1.5 cm]{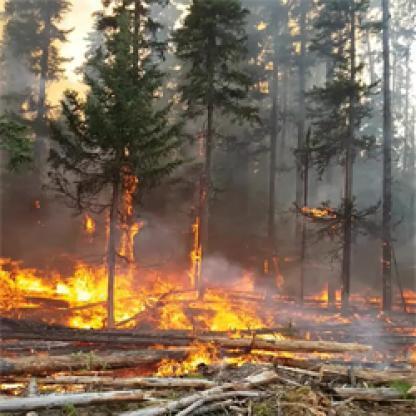} &
            \includegraphics[width=1.5 cm, height=1.5 cm]
            {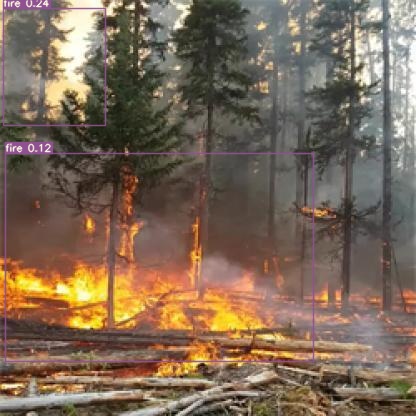} &
            \includegraphics[width=1.5 cm, height=1.5 cm]
            {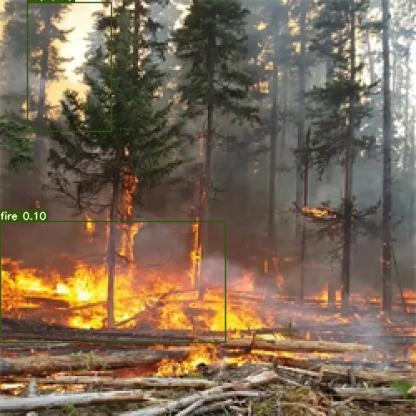} &
            \includegraphics[width=1.5 cm, height=1.5 cm]
            {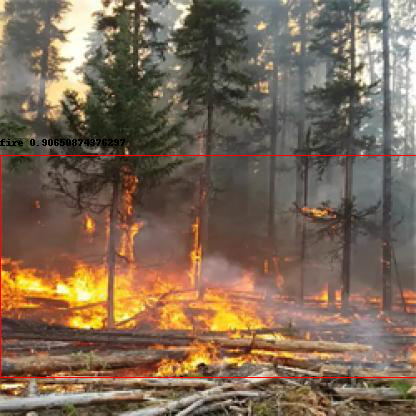} &
            \includegraphics[width=1.5 cm, height=1.5 cm]
            {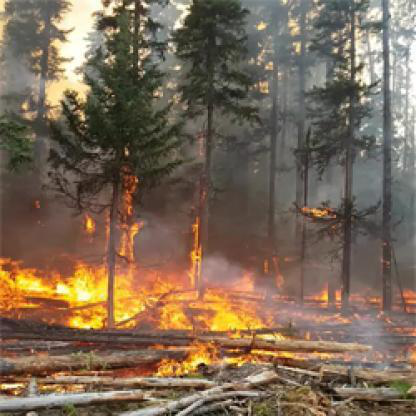} 
            \\
            \includegraphics[width=1.5 cm, height=1.5 cm]{sample_images/originals/rdr2_fire_274_jpg.rf.5cd62a192f168e339b5ce47d68f31f87} &
            \includegraphics[width=1.5 cm, height=1.5 cm]
            {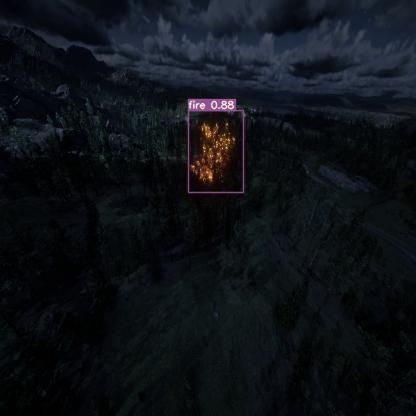} &
            \includegraphics[width=1.5 cm, height=1.5 cm]
            {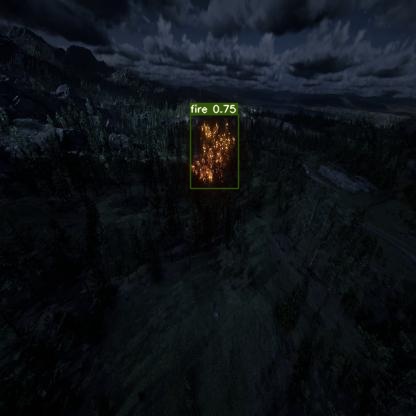} &
            \includegraphics[width=1.5 cm, height=1.5 cm]
            {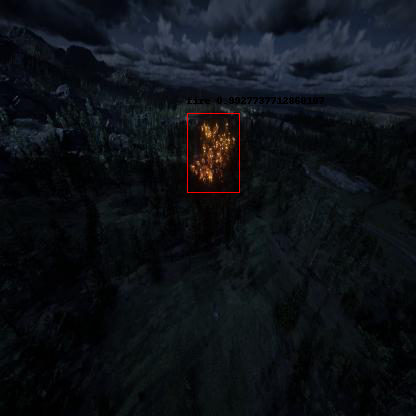} &
            \includegraphics[width=1.5 cm, height=1.5 cm]
            {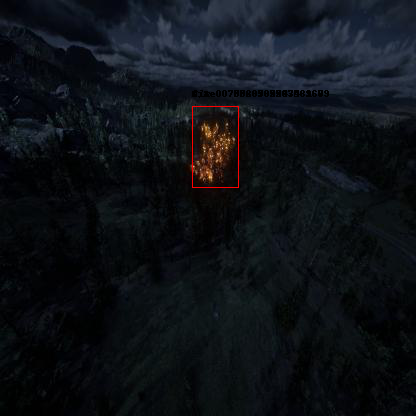} 
            \\
            \includegraphics[width=1.5 cm, height=1.5 cm]{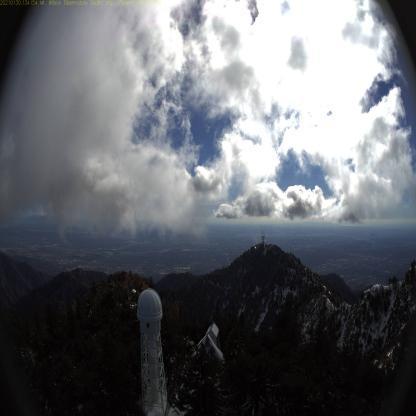} &
            \includegraphics[width=1.5 cm, height=1.5 cm]
            {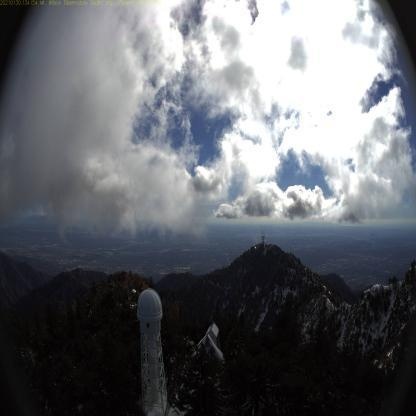} &
            \includegraphics[width=1.5 cm, height=1.5 cm]
            {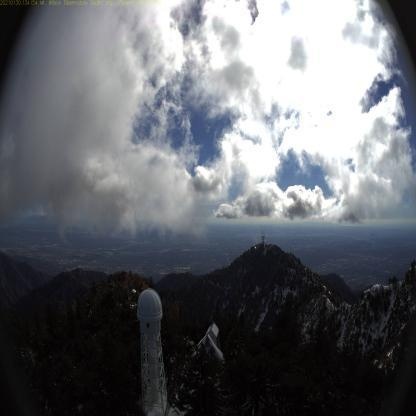} &
            \includegraphics[width=1.5 cm, height=1.5 cm]
            {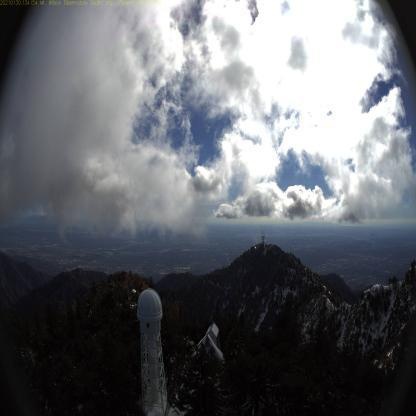} &
            \includegraphics[width=1.5 cm, height=1.5 cm]
            {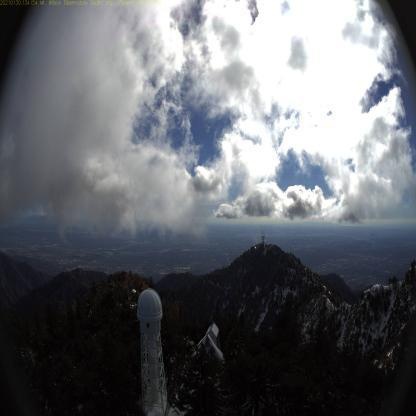} 
            \\
            \includegraphics[width=1.5 cm, height=1.5 cm]{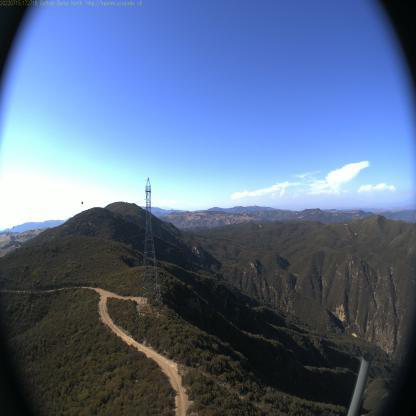} &
            \includegraphics[width=1.5 cm, height=1.5 cm]{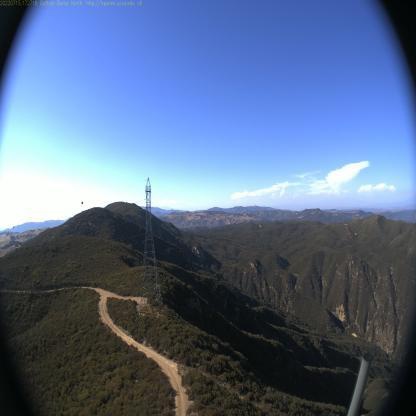} &
            \includegraphics[width=1.5 cm, height=1.5 cm]{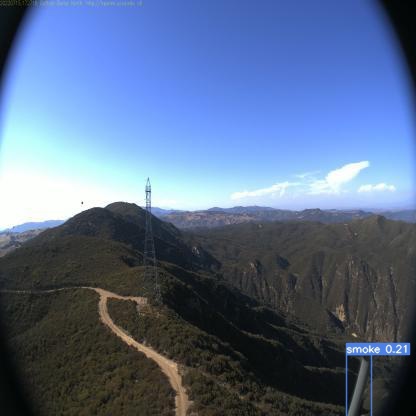} &
            \includegraphics[width=1.5 cm, height=1.5 cm]
            {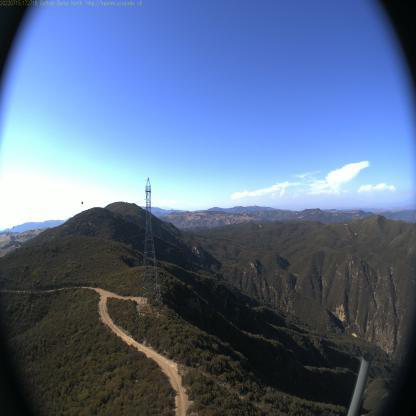} &
            \includegraphics[width=1.5 cm, height=1.5 cm]
            {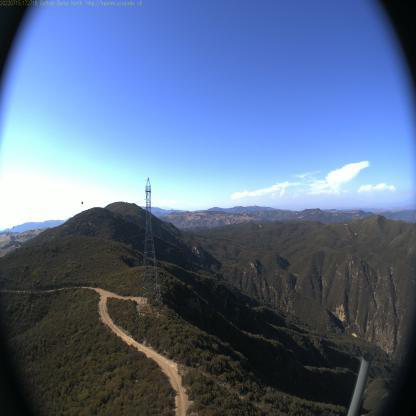} 
            \\
        \end{tabular}
        \caption{Inference results for localization.}
        \end{figure}

In this work we compared the performance of four object detectors namely YOLOv7~\cite{YOLOv7}, YOLOv7-tiny~\cite{YOLOv7}, Detection Transformer (DETR)~\cite{DETRECCV2022} and Deformable DETR~\cite{DefDETEICLR2021}. All these methods were pre-trained on COCO dataset~\cite{MSCOCO} and were fine tuned on the combined forest fire dataset that we proposed in this work. The implementation details of these methods is shown in Table~\ref{tab:traintest} and their comparative analysis is discussed in Table 5.

{\textbf{Classification Analysis}}: As part of our analysis, Deformable DETR was trained for 37 epochs with fewer model parameters, as compared to the original DETR model, which was trained for 65 epochs. As expected, it outperformed DETR with much higher accuracy, F1-score, and mAP value (see Tab.~\ref{tab:performance}, Fig.~\ref{fig:conf}) \\
In terms of performance, Deformable DETR has the highest mAP (mean Average Precision). Even though most of its predictions for alerts (i.e. smoke and fire warnings) are accurate, it has a low recall value of $0.65$ which means that about $35\%$ of the time, it cannot detect these warnings. As shown in Figure 3(a), the number of False Negatives is extremely high. \\ 
YOLOv7, compared to the rest of the models, gave us the highest number of false alerts. This means that it detects alerts in normal images. YOLOv7-tiny, which we trained using the same hyperparameters as YOLOv7, is the most accurate of the four models. It has achieved an accuracy of 88\%. \\ 
Comparing YOLOv7-tiny to Deformable DETR, we can see that the latter gives fewer false positives and has higher precision. However, YOLOv7-tiny reduces the number of missed fire and smoke alerts by half of that of Deformable DETR. However, it is important to note that results for Deformable DETR and DETR are computed at a higher confidence threshold of 0.5 as compared to Yolov7 models which are 0.1.\\

{\textbf{Localization Analysis}}: Localization results differ in bounding box size and confidence values. Deformable DETR gives bounding boxes with the highest confidence, as compared to other models. Yolov7 and Yolov7-tiny give boxes with some overlaps and a large range of confidence values. DETR, on the other hand, misses out on most smoke and fire.\\

{\textbf{Compute Cost Analysis Analysis}}: YOLOv7-tiny having the smallest number of parameters, gives us an inference time of $0.2$ ms as compared to that of $2.1$ ms of Deformable DETR. uses approx. $16\%$ of the parameters used in YOLOv7 hence reducing the need for heavyweight processors to run inference. This makes YOLOv7-tiny more accurate and faster predictions than any of our other models.

\section{Conclusion}
In this work we first filled the gap in the existing dataset by generating fire and smoke events synthetically via a game engine. We then combined our data with existing datasets and used it for performance evaluation of state-of-the-art algorithms such as YOLOv7, YOLOv7-tiny and variants of Detection Transformers. We for the first time performed forest fire prediction via transformers for which we introduced DETR~\cite{DETRECCV2022} and Deformable DETR~\cite{DefDETEICLR2021} for solving the problem of generating early warnings for wild fires. Low inference time and low missed detections are some of the important contraints associated with early warning systems. Since there is typically a human in the loop thus some amount of false postives can be accepted. 

Considering these constraints we found YOLOv7-tiny to be the most suited algorithm for the problem. Deformable DETR, on other hand, gives less false positives but at the cost of miss-detection. It also has a higher inference time and is computationally expensive due to its model complexity. YOLOv7, on the other hand, gives comparatively higher false positives, however, it has the lowest inference time and the highest accuracy out of the all models. In addition, it has a low memory footprint and can be executed on embedded devices such as Jetson Nano. It is also computationally light due to the low number of parameters which helps run it 24/7. In summary, out of the 4 models, YOLOv7-tiny is the best, most suited, with Deformable-DeTr as the runner-up. 


%
%
\scriptsize
\bibliographystyle{IEEEbib}
\bibliography{references}

@article{FoggiaTCSVT2015,
  title={Real-time fire detection for video-surveillance applications using a combination of experts based on color, shape, and motion},
  author={Foggia, Pasquale and Saggese, Alessia and Vento, Mario},
  journal={IEEE Transaction on circuits and systems for video technology},
  volume={25},
  number={9},
  pages={1545--1556},
  year={2015},
  publisher={IEEE}
}

@article{ren2015faster,
  title={Faster r-cnn: Towards real-time object detection with region proposal networks},
  author={Ren, Shaoqing and He, Kaiming and Girshick, Ross and Sun, Jian},
  journal={Advances in neural information processing systems},
  volume={28},
  year={2015}
}

@article{Analysiskukuk_comprehensive_2021,
	title = {Comprehensive {Analysis} of {Forest} {Fire} {Detection} using {Deep} {Learning} {Models} and {Conventional} {Machine} {Learning} {Algorithms}},
	issn = {2149-9144},
	url = {https://dergipark.org.tr/en/doi/10.22399/ijcesen.950045},
	doi = {10.22399/ijcesen.950045},
	urldate = {2022-03-28},
	journal = {International Journal of Computational and Experimental Science and Engineering},
	author = {Kukuk, Suha Berk and Kilimci, Zeynep Hilal},
	month = jul,
	year = {2021}
}

@article{MSERmatas2004robust,
  title={Robust wide-baseline stereo from maximally stable extremal regions},
  author={Matas, Jiri and Chum, Ondrej and Urban, Martin and Pajdla, Tom{\'a}s},
  journal={Image and vision computing},
  volume={22},
  number={10},
  pages={761--767},
  year={2004},
  publisher={Elsevier}
}

@article{ZHOUFire2016,
title = {Wildfire smoke detection based on local extremal region segmentation and surveillance},
journal = {Fire Safety Journal},
volume = {85},
pages = {50-58},
year = {2016},
issn = {0379-7112},
doi = {https://doi.org/10.1016/j.firesaf.2016.08.004},
url = {https://www.sciencedirect.com/science/article/pii/S0379711216301059},
author = {Zhiqiang Zhou and Yongsheng Shi and Zhifeng Gao and Sun Li}
}

@misc{rdr2_map, 
    title={Large detailed map of Red Dead Redemption 2 World},
    howpublished = {\url{https://www.mapsland.com/games/large-detailed-map-of-red-dead-redemption-2-world}}
}

@misc{ReddeadRedemption2,
  title = {Red Dead Redemption 2 - Cumberland Forest},
  howpublished = {\url{https://www.rockstargames.com/games/reddeadredemption2}},
  note = {Accessed: 2022-10-26}
}

@article{Saponara2021real,
  title={Real-time video fire/smoke detection based on CNN in antifire surveillance systems},
  author={Saponara, Sergio and Elhanashi, Abdussalam and Gagliardi, Alessio},
  journal={Journal of Real-Time Image Processing},
  volume={18},
  number={3},
  pages={889--900},
  year={2021},
  publisher={Springer}
}

@incollection{RobertCVBC2021fire,
  title={Fire Detection by Parallel Classification of Fire and Smoke Using Convolutional Neural Network},
  author={Robert Singh, A and Athisayamani, Suganya and Sankara Narayanan, S and Dhanasekaran, S},
  booktitle={Computational Vision and Bio-Inspired Computing},
  pages={95--105},
  year={2021},
  publisher={Springer}
}

@article{ThermalImagigPuLi2020,
title = {Image fire detection algorithms based on convolutional neural networks},
journal = {Case Studies in Thermal Engineering},
volume = {19},
pages = {100625},
year = {2020},
issn = {2214-157X},
doi = {https://doi.org/10.1016/j.csite.2020.100625},
url = {https://www.sciencedirect.com/science/article/pii/S2214157X2030085X},
author = {Pu Li and Wangda Zhao}
}

@inproceedings{DETRECCV2022,
  title={End-to-end object detection with transformers},
  author={Carion, Nicolas and Massa, Francisco and Synnaeve, Gabriel and Usunier, Nicolas and Kirillov, Alexander and Zagoruyko, Sergey},
  booktitle={European Conference on Computer Vision},
  pages={213--229},
  year={2020},
  organization={Springer}
}

@TECHREPORT{NOAAClimate2022,
  TITLE =         {State of the Climate: Monthly Global Climate Report for Annual 2021},
  INSTITUTION =   {NOAA National Centers for Environmental Information},
  MONTH =         {January},
  YEAR  =         {2022},

}

@article{RothermelVacchiano2014,
author = {Vacchiano, Giorgio and Ascoli, Davide},
year = {2014},
month = {01},
pages = {},
title = {An Implementation of the Rothermel Fire Spread Model in the R Programming Language},
volume = {51},
journal = {Fire Technology},
doi = {10.1007/s10694-014-0405-6}
}

@article{Yang_Lupascu_Meel_2021, 
title={Predicting Forest Fire Using Remote Sensing Data And Machine Learning}, 
author={Yang, Suwei and Lupascu, Massimo and Meel, Kuldeep S.}, 
volume={35}, url={https://ojs.aaai.org/index.php/AAAI/article/view/17758}, 
DOI={10.1609/aaai.v35i17.17758}, 
number={17}, 
journal={Proceedings of the AAAI Conference on Artificial Intelligence}, 
year={2021}, 
month={May}, 
pages={14983-14990} }

@article{PengCEA2019, 
title={Real-time forest smoke detection using hand-designed features and deep learning}, 
author={Peng, Yingshu and Wang, Yi},
volume={167}, 
issue={C}, 
journal={Computers and Electronics in Agriculture}, 
year={2019}, 
month={Dec}}

@INPROCEEDINGS{JosephYoloCVPR2017,  
author={Redmon, Joseph and Farhadi, Ali},  
booktitle={IEEE Conference on Computer Vision and Pattern Recognition},   title={YOLO9000: Better, Faster, Stronger},   
year={2017}}

@Article{QingMDPI2022,
AUTHOR = {An, Qing and Chen, Xijiang and Zhang, Junqian and Shi, Ruizhe and Yang, Yuanjun and Huang, Wei},
TITLE = {A Robust Fire Detection Model via Convolution Neural Networks for Intelligent Robot Vision Sensing},
JOURNAL = {Sensors},
VOLUME = {22},
YEAR = {2022},
NUMBER = {8},
ARTICLE-NUMBER = {2929},
URL = {https://www.mdpi.com/1424-8220/22/8/2929},
PubMedID = {35458913},
ISSN = {1424-8220},
}

@misc{aiformankind, title={AIFORMANKIND/wildfire-smoke-dataset: Open wildfire smoke datasets}, howpublished = {\url{https://github.com/aiformankind/wildfire-smoke-dataset}}, journal={GitHub}, author={Aiformankind}}

@misc{alik05_2022, title={Forest fire dataset},
 howpublished = {\url{https://www.kaggle.com/datasets/alik05/forest-fire-dataset}}, journal={Kaggle}, author={Alik05}, year={2022}, month={Apr}}

@misc{hpwren, title={Cameras from various HPWREN related sites},
 howpublished = {\url{http://hpwren.ucsd.edu/cameras/}}, journal={HPWREN}}

@misc{khan_2020, title={Dataset for forest fire detection}, howpublished = {\url{https://data.mendeley.com/datasets/gjmr63rz2r/1}}, journal={Mendeley Data}, publisher={Mendeley Data}, author={Khan, Ali and Hassan, Bilal}, year={2020}, month={Aug}}

@misc{kutlu_2021, title={Forest fire},
 howpublished = {\url{https://www.kaggle.com/datasets/kutaykutlu/forest-fire}}, journal={Kaggle}, author={Kutlu, Kutay}, year={2021}, month={Mar}}

@misc{mankind_2022, title={Wildfire smoke object detection dataset - raw},howpublished = {\url{https://public.roboflow.com/object-detection/wildfire-smoke/1/download}}, journal={Roboflow}, author={Mankind, AI For}, year={2022}, month={Aug}}

@misc{mivia,
  title        = {Fire detection dataset},
  howpublished = {\url{https://mivia.unisa.it/datasets/video-analysis-datasets/fire-detection-dataset/}},
  journal      = {MIVIA},
  author       = {Mivia}
}

@misc{mivia_smoke, title={Smoke detection dataset},
 howpublished = {\url{https://mivia.unisa.it/datasets/video-analysis-datasets/smoke-detection-dataset/}}, journal={MIVIA}, author={Mivia}}

@misc{nist_2021,
  title        = {Fire},
  howpublished = {\url{https://www.nist.gov/fire}},
  organization = {National Institute of Standards and Technology (NIST)},
  year         = {2021},
  month        = apr
}

@misc{imagescvdataset,
  title        = {Image Datasets for Computer Vision and Machine Learning},
  howpublished = {\url{https://images.cv/search-labeled-image-dataset}},
  organization = {Images.CV},
  year         = {2023}
}

@inproceedings{DefDETEICLR2021,
  author    = {Xizhou Zhu and
               Weijie Su and
               Lewei Lu and
               Bin Li and
               Xiaogang Wang and
               Jifeng Dai},
  title     = {Deformable {DETR:} Deformable Transformers for End-to-End Object Detection},
  booktitle = {9th International Conference on Learning Representations, {ICLR} 2021,
               Virtual Event, Austria, May 3-7, 2021},
  publisher = {OpenReview.net},
  year      = {2021},
  url       = {https://openreview.net/forum?id=gZ9hCDWe6ke},
  bibsource = {dblp computer science bibliography, https://dblp.org}
}

@article{YOLOv7,
  author    = {Wang, Chien-Yao and Bochkovskiy, Alexey and Liao, Hong-Yuan Mark},
  title     = {YOLOv7: Trainable Bag-of-Freebies Sets New State-of-the-Art for Real-Time Object Detectors},
  journal   = {arXiv preprint arXiv:2207.02696},
  year      = {2022},
  url       = {https://arxiv.org/abs/2207.02696}
}

@InProceedings{MSCOCO,
author="Lin, Tsung-Yi
and Maire, Michael
and Belongie, Serge
and Hays, James
and Perona, Pietro
and Ramanan, Deva
and Doll{\'a}r, Piotr
and Zitnick, C. Lawrence",
editor="Fleet, David
and Pajdla, Tomas
and Schiele, Bernt
and Tuytelaars, Tinne",
title="Microsoft COCO: Common Objects in Context",
booktitle="Computer Vision -- ECCV 2014",
year="2014",
publisher="Springer International Publishing",
address="Cham",
pages="740--755"
}

\end{document}